\documentclass[letterpaper, 10 pt, conference]{ieeeconf}  

\IEEEoverridecommandlockouts                              

\usepackage{booktabs, makecell, threeparttable}
\usepackage{array}
\usepackage{multirow} 
\makeatletter
\let\NAT@parse\undefined
\makeatother
\usepackage[colorlinks,citecolor=blue,urlcolor=blue,bookmarks=false,hypertexnames=true]{hyperref}
\usepackage[numbers,sort&compress]{natbib}
\usepackage{graphicx}
\usepackage{url}

\usepackage{algorithm}
\usepackage{times}
\usepackage{epsfig}
\usepackage{amsmath}
\usepackage{amssymb}
\usepackage{algpseudocode}
\usepackage{amsfonts,bm}

\usepackage{booktabs}
\usepackage{bbding}
\usepackage{epigraph}
\usepackage{tcolorbox}
\usepackage{colortbl, xcolor}

\usepackage{graphicx}
\usepackage{titlesec}

\newcommand{\logopic}{\includegraphics[height=3.5ex]{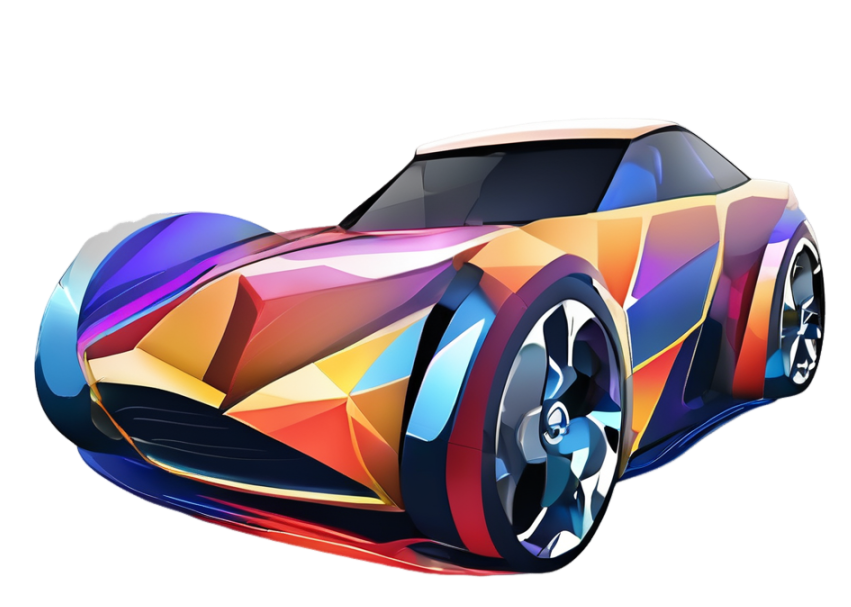}}

\newcommand{\logopicsmall}{\includegraphics[height=2ex]{logo.png}}

\definecolor{grey}{rgb}{0.1,0.1,0.1} 
\usepackage{pifont}

\title{\LARGE \bf \logopic DRIVE: Dependable Robust Interpretable Visionary Ensemble Framework in Autonomous Driving}

\author{Songning Lai$^{*1}$, Tianlang Xue$^{*1}$, Hongru Xiao$^{2}$, Lijie Hu$^{3}$, Jiemin Wu$^{1}$, Ninghui Feng$^{1}$, Runwei Guan$^{1}$, \\Haicheng Liao$^{4}$, Zhenning Li$^{4}$, Yutao Yue$^{\dagger1}$ \\
\href{https://xll0328.github.io/project/DRIVE/}{Project Website: https://xll0328.github.io/project/DRIVE/}
\thanks{$^{*}$ Contributions equally.}
\thanks{$^{\dagger}$ Correspondence to Yutao Yue \{yutaoyue@hkust-gz.edu.cn\}.}
\thanks{$^{1}$ The Hong Kong University of Science and Technology (Guangzhou)}%
\thanks{$^{2}$ Tongji University}%
\thanks{$^{3}$ King Abdullah University of Science and Technology}%
\thanks{$^{4}$ University of Macau}
\thanks{This work was supported by Guangzhou-HKUST(GZ) Joint Funding Program (Grant No.2023A03J0008), Education Bureau of Guangzhou Municipality.}
\vspace{-10pt}}

\begin{document}

\maketitle
\thispagestyle{empty}
\pagestyle{empty}

\begin{abstract}

Recent advancements in autonomous driving have seen a paradigm shift towards end-to-end learning paradigms, which map sensory inputs directly to driving actions, thereby enhancing the robustness and adaptability of autonomous vehicles. However, these models often sacrifice interpretability, posing significant challenges to trust, safety, and regulatory compliance. To address these issues, we introduce \logopicsmall\underline{\textbf{DRIVE}} -- \underline{\textbf{D}}ependable \underline{\textbf{R}}obust \underline{\textbf{I}}nterpretable \underline{\textbf{V}}isionary \underline{\textbf{E}}nsemble Framework in Autonomous Driving, a comprehensive framework designed to improve the dependability and stability of explanations in end-to-end unsupervised autonomous driving models. Our work specifically targets the inherent instability problems observed in the Driving through the Concept Gridlock (DCG) model, which undermine the trustworthiness of its explanations and decision-making processes. We define four key attributes of \textbf{DRIVE}: consistent interpretability, stable interpretability, consistent output, and stable output. These attributes collectively ensure that explanations remain reliable and robust across different scenarios and perturbations. Through extensive empirical evaluations, we demonstrate the effectiveness of our framework in enhancing the stability and dependability of explanations, thereby addressing the limitations of current models. Our contributions include an in-depth analysis of the dependability issues within the DCG model, a rigorous definition of \textbf{DRIVE} with its fundamental properties, a framework to implement \textbf{DRIVE}, and novel metrics for evaluating the dependability of concept-based explainable autonomous driving models. These advancements lay the groundwork for the development of more reliable and trusted autonomous driving systems, paving the way for their broader acceptance and deployment in real-world applications.

\end{abstract}

\vspace{-5pt}
\epigraph{``We can only see a short distance ahead, but we can see plenty there that needs to be done.'' - {Alan Turing}}{}
\vspace{-12pt}

\section{INTRODUCTION}

\begin{figure*}[t]
\centering
\includegraphics[width=1.6\columnwidth]{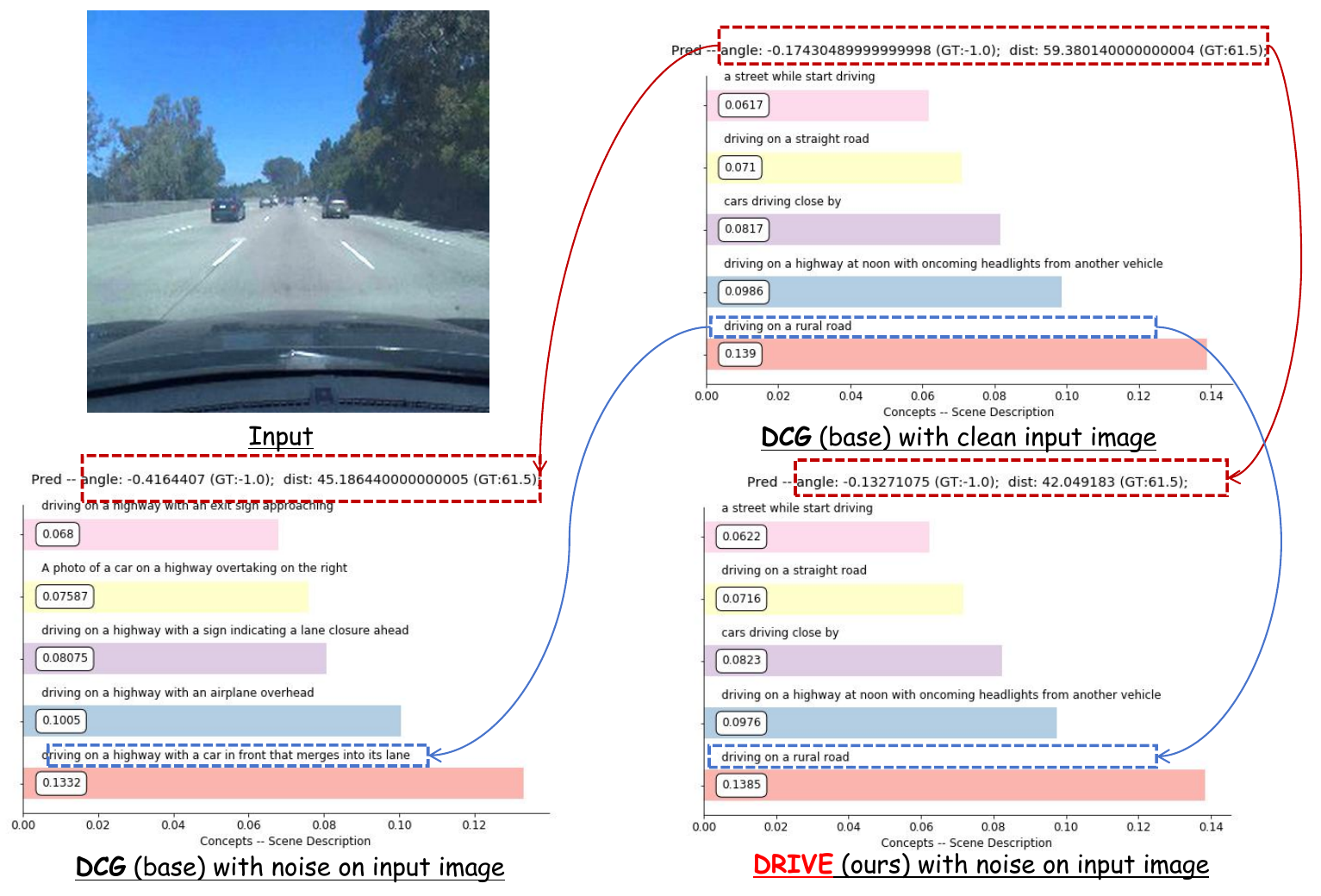}
  \vspace{-8pt}
  \caption{Top right and bottom left figures show that the interpretable and predicted outputs of DCG are sensitive to perturbations, and on the contrary, our optimization framework DRIVE (bottom right) can cope with this very well.}
  \label{fig:intro}
  \vspace{-20pt}
\end{figure*}

Autonomous driving has seen rapid advancements in recent years, driven by innovations in deep learning and artificial intelligence \cite{yurtsever2020survey,zhao2023autonomous}. However, despite these technological strides, a notable lack of public trust persists, primarily due to feelings of loss of control and the opacity of decision-making processes within these systems. For autonomous vehicles to achieve widespread commercial adoption, it is crucial to foster public acceptance and confidence. A key challenge in achieving this is the development of models that provide robust and interpretable explanations of their behaviors \cite{li2023trustworthy}.

Currently, many autonomous driving models based on deep learning operate as black boxes, lacking transparency which hinders understanding and trust. Previous approaches have attempted to address this issue through post-hoc explanations, such as those generated by GPT models \cite{liao2024gpt,liao2024cognitive}, but these methods fall short because they offer justifications after the fact, rather than providing clear insights into the decision-making process as it unfolds \cite{molnar2020interpretable,wen2023road}.

Anticipatory or prospective explanations, which aim to clarify the reasoning behind decisions before they are made, could significantly enhance trustworthiness. However, achieving this level of interpretability presents numerous challenges. To address these issues, we explored the Driving through the Concept Gridlock (DCG) model \cite{wang2024drive}(See Section \ref{sec:DCG} for details), a post-hoc explainable autonomous driving framework designed to enhance the interpretability of deep learning models \cite{liao2024human}. The DCG model offers several advantages, including its ability to generate detailed explanations of the decision-making process, thereby increasing the transparency and trustworthiness of autonomous driving systems \cite{liao2024bat}. This framework has demonstrated effectiveness in providing clear insights into how decisions are made, making it a valuable tool for understanding the inner workings of complex models.

Although the advanced DCG model in explainable autonomous driving excels in prediction, our research has identified dependability issues. Specifically, the model exhibits sensitivity to input perturbations, differences in concept space, and variations in parameters (See in Figure \ref{fig:intro}), which lead to inconsistent behavior and unstable performance. This instability makes the reasoning process of the DCG model opaque, thereby undermining its credibility. This instability is particularly problematic in scenarios involving unlabelled data and self-supervised training, where the model's reasoning becomes opaque and its credibility is undermined \cite{rudin2019stop}.

To illustrate, the European Union (EU) has taken steps to ensure transparency in autonomous systems, such as the introduction of the General Data Protection Regulation (GDPR) \cite{gdpr2016general}, which includes provisions for the right to explanation, granting individuals the ability to seek clarity on decisions made by such systems. Additionally, Article 22 of the GDPR sets forth guidelines concerning the rights and responsibilities of stakeholders when employing automated decision-making technologies \cite{gdpr2022automated}. Therefore, developing stable and interpretable autonomous driving models capable of providing consistent and reliable explanations is not only a foundational step towards ensuring safe integration but also a critical aspect of meeting regulatory expectations and building public trust \cite{omeiza2021explanations,atakishiyev2024explainable}.

To address these challenges, we introduce the concept of \underline{\textbf{DRIVE}} -- \underline{\textbf{D}}ependable \underline{\textbf{R}}obust \underline{\textbf{I}}nterpretable \underline{\textbf{V}}isionary \underline{\textbf{E}}nsemble Framework in Autonomous Driving, which ideally possesses four key attributes within the same concept space:

(i) \underline{Consistent Interpretability}: There should be a significant overlap between the top-$k$ indices of "DRIVE" and the original DCG's explainable outputs, ensuring the consistent of explanations.

(ii) \underline{Stable Interpretability}: The explainable outputs should be stable to noise and perturbations.

(iii) \underline{Consistent Output}: The prediction distribution should closely resemble the original DCG's output, maintaining its superior performance.

(iv) \underline{Stable Output}: The distribution of outputs should remain robust during self-supervised learning and LLM concept set generation, even under disturbances.

Based on this definition of DRIVE, we devise a framework to find a DRIVE. Our empirical evaluations demonstrate the effectiveness of our framework. Our contributions are as follows:

\begin{itemize}

\item \textbf{In-depth Analysis of DCG Dependability.} We identify a significant dependability issue in generating textual descriptions from visual data due to alignment instability between text and images.

\item \textbf{Rigorous Definition of DRIVE.} Based on our findings, we formally define DRIVE and its four fundamental properties, offering a clear framework for evaluating explainable autonomous driving.

\item \textbf{A Framework to Find a DRIVE.} To address DCG fidelity issues, we propose a framework that optimizes parameters with minimal alteration, preserving the integrity of pre-trained large model parameters and aligning with our dependability definition. Our experiments have validated the effectiveness of this framework.

\end{itemize}

\section{RELATED WORK}

\noindent\textbf{Explainable Autonomous Driving. }End-to-end approaches in autonomous driving show significant potential but are criticized for their opaque decision-making, lacking transparency and interpretability. This opacity hinders debugging, reliability, and public trust \cite{2022Explainability}. Researchers have explored methodologies to enhance interpretability, such as attention mechanisms to highlight critical areas \cite{8237582}, intermediate attention maps to focus on relevant driving tasks \cite{9710855}, and analyzing object targets to reveal decision-making intentions \cite{9157111}. Zeng et al. \cite{8954347} developed a neural motion planner that generates interpretable representations, and Echterhoff et al. \cite{Echterhoff_2024_WACV} introduced the Concept Bottleneck Model (CBM) for linking decisions to specific objects. Real-world driving conditions, marked by environmental and regional variability, introduce noise and interference, challenging model interpretability \cite{10287869,10009752,Almalioglu2022}. Creating conceptually interpretable annotations becomes costly, and unsupervised methods may yield unstable results \cite{oikarinenlabel,lai2023faithful}. This paper aims to enhance the robustness of interpretable models in diverse scenarios, focusing on maintaining interpretability through intermediate outputs under challenging conditions.

\noindent\textbf{Dependability for Explainable Method. }Dependability in explicable methodologies involves reliably reflecting the true reasoning process, encompassing principles like sensitivity, implementation invariance, input invariance, and completeness \cite{hu2023seat,wiegreffe2019attention,jacovi2020towards,lyu2022towards,yeh2019fidelity}. Completeness ensures all relevant factors are captured, while stability under various perturbations ensures adaptability to significant changes and robustness against minor perturbations \cite{adebayo2018sanity,yin2022sensitivity,lai2024fts}. Initial strategies, such as smoothing techniques \cite{yeh2019fidelity} and iterative gradient descent algorithms \cite{yin2022sensitivity}, have aimed to achieve stable interpretations. In autonomous driving, post-hoc enhancements fail to provide reliable real-time decision support, as accidents can be sudden and initial deviations may lead to irreversible outcomes \cite{Dodge_2019,DING2024120057}. Therefore, this paper proposes a framework ensuring stable, real-time interpretability for end-to-end decision support. This method aims to enhance real-time interpretability and improve decision-making robustness in perturbed environments, increasing the applicability of autonomous driving algorithms.

\section{Preliminary-DCG}
\label{sec:DCG}

This section outlines the methodology employed in developing concept bottleneck models for automated driving (AD) applications named DCG, focusing on predicting a target value $y \in \mathbb{R}$ from an input $x \in \mathbb{R}^d$, while also providing reasoning $c \in \mathbb{R}^k$ for the prediction.

\noindent\textbf{Problem Formulation. }Given a set of training examples $\{(x^{(i)}, y^{(i)})\}_{i=1}^n$, the goal is to predict $y^{(i)}$ and its associated reasoning $c^{(i)}$ using a model of the form $f(g(x))$. Here, $g: \mathbb{R}^d \to \mathbb{R}^k$ maps the input into a concept space, and $f: \mathbb{R}^k \to \mathbb{R}$ maps the concepts to the final prediction. These models are referred to as concept bottleneck models \cite{koh2020concept, oikarinen2023label}.

\noindent\textbf{Feature Extraction. }The feature extraction component, denoted by $g$, plays a critical role in mapping the input $x$ into a meaningful concept space. In this previous work, a pre-trained concept-aware backbone inspired by video vision transformers \cite{neimark2021video, vivit} is utilized, which extracts features $F_{\text{input}}$ for further processing.

\noindent\textbf{Concept Bottleneck. }To incorporate explainability, a concept bottleneck is introduced. Driving scenarios $s$ are constructed to describe scenes and encode contextual information, such as road conditions, traffic density, and weather. These scenarios are generated using the generative capabilities of GPT-3.5 \cite{ouyang2022training} and complemented with human-created scene descriptions from the NuScenes dataset \cite{caesar2020nuscenes}, following the template of \emph{a photo of...} \cite{radford2021learning}. 

For each scenario, an image encoder $g_{\text{image}}: \mathbb{R}^d \to \mathbb{R}^l$ and a text encoder $g_{\text{text}}: \mathbb{R}^s \to \mathbb{R}^l$ are used to embed the images and scenarios, respectively. The similarity between the image embeddings and the scenario embeddings is measured using cosine similarity:

\vspace{-15pt}
\begin{equation}
\text{sim}_{\text{cos}}(x,s) = \frac{{g_{\text{image}}(x) \cdot g_{\text{text}}(s)}}{{\|g_{\text{image}}(x)\|_2 \|g_{\text{text}}(s)\|_2}}
\end{equation}
\vspace{-10pt}

\noindent where $\cdot$ denotes the dot product and $\|\cdot\|_2$ is the Euclidean norm.

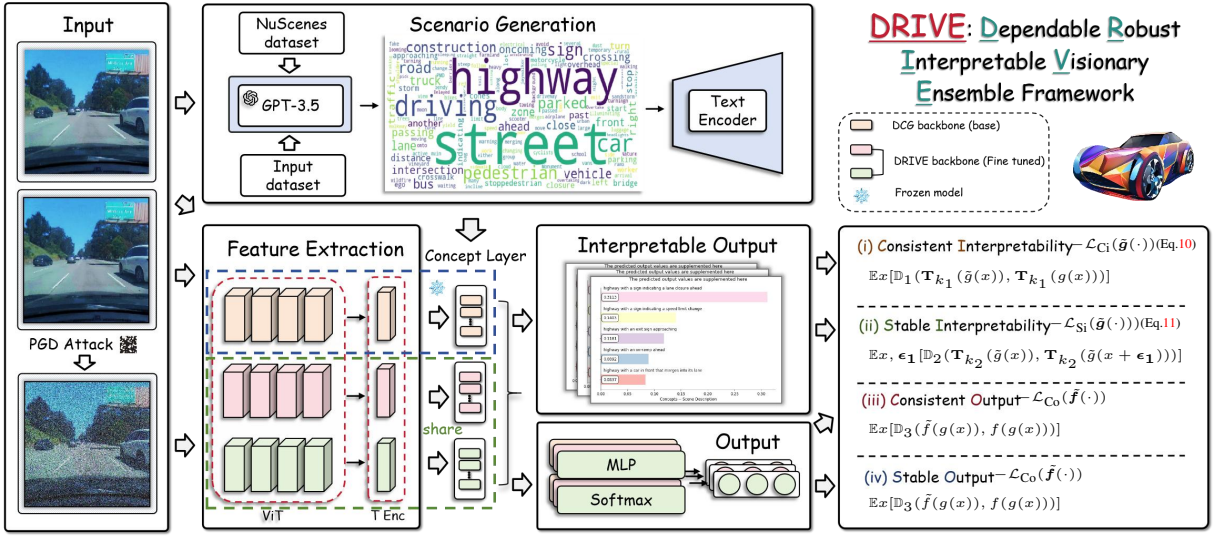
\begin{figure*}[t]
\centering
\includegraphics[width=1.9\columnwidth]{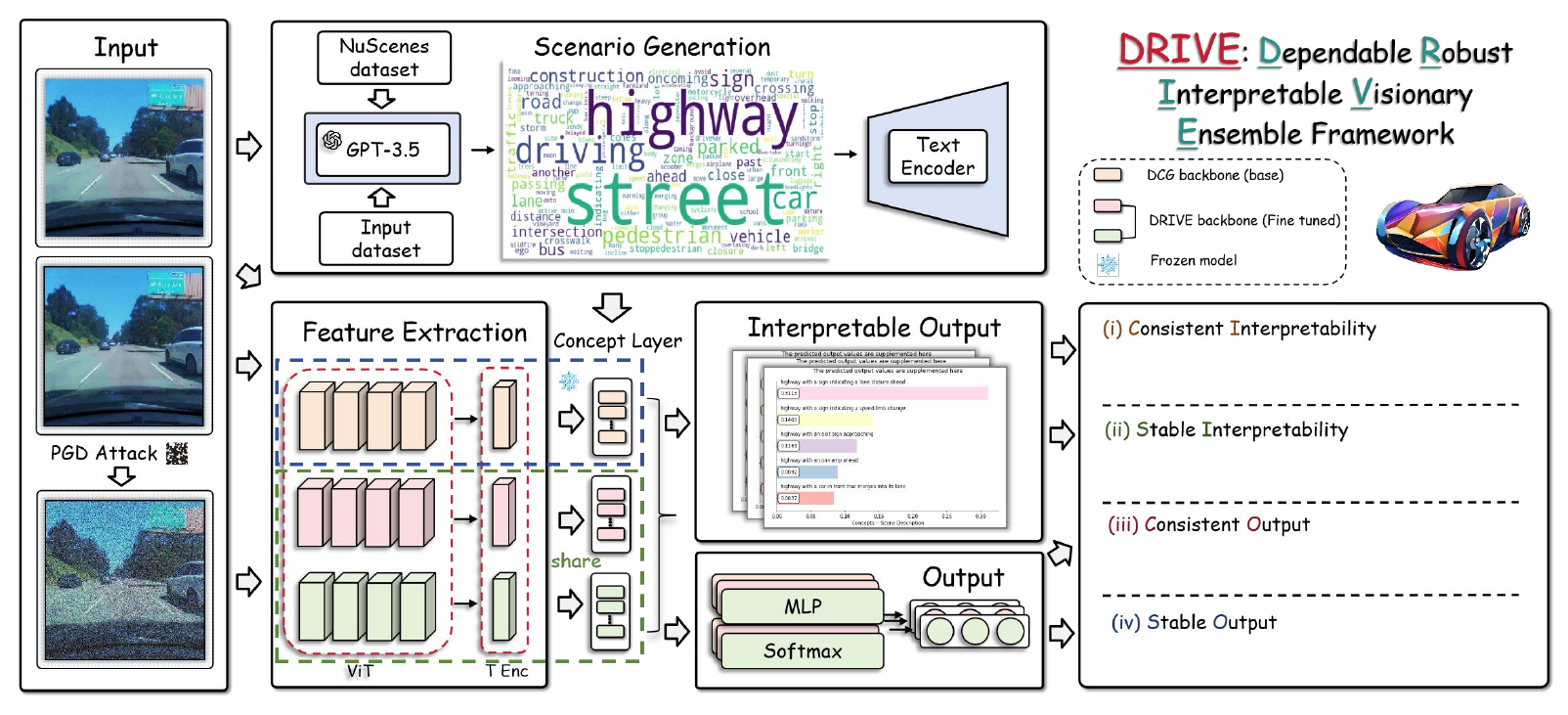}
		\put(-58,114){\tiny$-\mathcal{L}_{\text{Ci}}(\Tilde{\bm{g}}(\cdot))$(Eq.\ref{eq:lcin})}
  	\put(-134,102){\tiny$\mathbb{E}x [\mathbb{D}_1(\mathbf{T}_{k_1}(\Tilde{g}(x)),\mathbf{T}_{k_1}(g(x)))]$}
		\put(-66,84){\tiny$-\mathcal{L}_{\text{Si}}(\Tilde{\bm{g}}(\cdot)))$(Eq.\ref{eq:sin})}
  	\put(-134,72){\tiny$\mathbb{E}{x, \bm{\epsilon_1}} [\mathbb{D}_2(\mathbf{T}_{k_2}(\Tilde{g}(x)), \mathbf{T}_{k_2}(\Tilde{g}(x + \bm{\epsilon_1})))]$}
   		\put(-78,56){\tiny$-\mathcal{L}_{\text{Co}}(\Tilde{\bm{f}}(\cdot))$}
  	\put(-134,44){\tiny$\mathbb{E}x [\mathbb{D}_3(\Tilde{f}(g(x)), f(g(x)))]$}
   		\put(-87,27){\tiny$-\mathcal{L}_{\text{Co}}(\Tilde{\bm{f}}(\cdot))$}
  	\put(-134,16){\tiny$\mathbb{E}x [\mathbb{D}_3(\Tilde{f}(g(x)), f(g(x)))]$}
  \vspace{-8pt}
  \caption{Overall pipeline of DRIVE. The input is processed by a feature extractor and a temporal encoder, followed by a concept bottleneck with scenario encoding. The DRIVE model incorporates a multi-objective optimization process, balancing consistent interpretability (Ci), stable interpretability (Si), consistent output (Co), and stable output (So) through auxiliary loss functions. The model is trained using PGD to enhance robustness against perturbations while maintaining interpretability and predictive consistency.}
  \label{fig:model}
  \vspace{-20pt}
\end{figure*}

\noindent\textbf{Temporal Encoding. }The temporal encoding component employs a Longformer architecture \cite{beltagy2020longformer} to capture the temporal dynamics of the input sequences. Given a sequence of features $F_{\text{input}}$ and a sliding window size $w$, the attention mechanism is defined as:

\vspace{-15pt}
\begin{equation}
\text{Attention}(Q, K, V) = \text{softmax}\left(\frac{QK^\top}{\sqrt{d_k}}\right)V
\end{equation}
\vspace{-10pt}

\noindent where $Q$, $K$, and $V$ are the query, key, and value matrices, respectively, and $d_k$ is the dimensionality of the key vectors. The Longformer's sliding window attention reduces the computational complexity from $O(n^2)$ to $O(w \times n)$.

A special token ([CLS]) is prepended to the sequence, and the Longformer maintains global attention on this token. The final state of the features related to the [CLS] token is used as the representation of the video. This representation is then passed through a multi-layer perceptron (MLP) head for regression tasks, consisting of two linear layers with a GELU activation function \cite{DBLP:journals/corr/HendrycksG16} and dropout \cite{srivastava2014dropout}.

\noindent\textbf{Training. }The models are trained using the root mean squared error (RMSE) loss:

\vspace{-10pt}
\begin{equation}
\mathcal{L}_{init} = \sqrt{\frac{\sum_{i=1}^N(f(g(x^{(i)})) - y^{(i)})^2}{N}}
\end{equation}
\vspace{-10pt}

\noindent where $g(x^{(i)}) = \textit{sim}(x^{(i)},s)$, and $N$ denotes the number of training examples.

Despite its advanced predictive capabilities, the DCG model exhibits vulnerabilities such as sensitivity to input perturbations, parameter variations and the difference of concept spaces, leading to inconsistent performance. These shortcomings raise concerns about its reliability and suitability for real-world scenarios where dependability is paramount. Specifically, the reliance on GPT-3.5 for generating driving scenarios may introduce inaccuracies, and the use of pre-trained models for the concept bottleneck, especially in an unsupervised or self-supervised manner, may not fully capture the nuanced relationships between driving concepts and their visual representations. Addressing these limitations is crucial for enhancing the trustworthiness of vision-based explainable AD systems.

\section{ANALYSIS AND DEFINATION}

\noindent\textbf{Dependability Issues in DCG.} The state-of-the-art DCG model in explainable autonomous driving excels in prediction but shows sensitivity to input perturbations, concept space differences, and parameter variations, leading to inconsistent performance. These issues undermine its reliability for real-world applications. Our focus is on stabilizing DCG to enhance trust in vision-based explainable driving systems. Addressing these challenges will contribute to developing more dependable and resilient models, boosting confidence in their predictive accuracy across various applications.

\noindent\textbf{What is ``Dependably Explainable Automated Driving"?} "Dependably Explainable Automated Driving" denotes an AD system that offers clear, understandable explanations for its decisions and ensures reliable performance across various scenarios. By providing transparent and interpretable decision-making, this concept enhances trust and promotes safety, emphasizing reliability, transparency, and interpretability in the deployment of AD technologies.

\noindent{\textbf{\underline{Definition 1} for Top-$k$ Overlaps.}
This definition quantifies the overlap of the top-$k$ concept weights to measure the similarity between concepts. For a vector \( x \in \mathbb{R}^d \), we define the set of top-$k$ components \( T_k(\cdot) \):

\begin{equation*}
    T_k(x) = \{ i : i \in [d] \text{ and } |\{ j : x_j \geq x_i \text{ and } j \in [d] \}| \leq k \}.
\end{equation*}
Given two vectors \( x \) and \( x' \), the top-$k$ overlap function \( V_k(x, x') \) is defined as the ratio of the intersection of their top-$k$ components:

\vspace{-13pt}
\begin{equation*}
    V_k(x, x') = \frac{1}{k} | T_k(x) \cap T_k(x') |.
\end{equation*}
\vspace{-10pt}

\noindent\textbf{\underline{DEFINATION 2} for a Dependable Robust Interpretable Visionary Ensemble (\underline{DRIVE}) Model}
\label{sec:def_drive}

A DRIVE model is designated as $\bm{(\gamma_i,\epsilon_j,\rho_q,\mathbb{D}_k)}$ where $i,k \in \{1,2,3,4\}$ and $j,q \in \{1,2\}$ - Dependable if it meets the following criteria for any input $x \in \mathbb{R}^d$:

\noindent \textbf{(i) Consistent Interpretability (Ci)}: 
      The interpretability of the model's output should be within a bounded distance from that of its original counterpart. Formally, $\mathbb{D}_1(\Tilde{g}(x), g(x) \leq \gamma_1$  for some non-negative threshold $\gamma_1 \geq 0$. Here $\mathbb{D}_1$ denotes a is measure of probabilistic divergence, $g(x)$ is the interpretable ouput of the baseline DCG and $\Tilde{g(x)}$ is the corresponding output of the DRIVE model.
    
\noindent \textbf{(ii) Stable Interpretability (Si)}: 
    The interpretability must remain stable under perturbations to the input. Specifically, $\mathbb{D}_2(\Tilde{g}(x), \Tilde{g}(x + \bm{\epsilon_1})) \leq \gamma_2$ for all perturbations $\|\bm{\epsilon_1}\| \leq \rho_1$. In this context, $\mathbb{D}_2$ is a measure of probabilistic divergence, $\|\cdot\|$ is a norm, and $\rho_1 \geq 0$ sets the maximum allowable perturbation magnitude.

\noindent \textbf{(iii) Consistent Output (Co)}: 
The output of the model should closely match that of the original DCG. This is quantified as $\mathbb{D}_3(\Tilde{f}(g(x)), f(g(x))) \leq \gamma_3$ for some non-negative threshold $\gamma_3 \geq 0$. Here, $\mathbb{D}_3$ is a measure of probabilistic divergence, $f(g(x))$ represents the predictive output of the baseline DCG, and $\Tilde{f}(g(x))$ is the output of the proposed DRIVE model.

\noindent \textbf{(iv) Stable Output (So)}: The predictive output should also exhibit stability when subjected to input perturbations. Thus, $\mathbb{D}_4(\Tilde{f}(g(x)), \Tilde{f}(g(x + \bm{\epsilon_2}))) \leq \gamma_4$ for all perturbations $\|\bm{\epsilon_2}\| \leq \rho_2$. In this case, $\mathbb{D}_4$ is a measure of probabilistic divergence, $\|\cdot\|$ is a norm, and $\rho_2 \geq 0$ defines the maximum perturbation magnitude allowed.

A DRIVE model, defined by its unwavering reliability, distinguishes itself through consistent predictive accuracy, robust interpretability, and stable performance despite variations in input data or internal model dynamics. 

\vspace{-5pt}
\section{Constructing \logopicsmall DRIVE}
\label{CAD}
\vspace{-5pt}

In order to construct a $\bm{(\gamma_i,\epsilon_j,\rho_q,\mathbb{D}_k)}$ - DRIVE, we formulate a multi-objective optimization problem that aligns with the attributes defined in Section \ref{sec:def_drive}. This formulation leads to a comprehensive objective function that balances the four critical aspects of the model: consistent interpretability (Ci), stable interpretability (Si), consistent output (Co), and stable output (So). The framework is depicted in Figure \ref{fig:model}.

Given the definitions of Ci, Si, Co, and So, we aim to find a set of model parameters ${\Tilde{f}(\cdot),\Tilde{g}(\cdot)}$ that satisfy the following optimization problem. Formally,

\vspace{-15pt}
\begin{align}
\min [\mathcal{L}_{\text{Ci}}(\Tilde{\bm{g}}(\cdot)) + \mathcal{L}_{\text{Si}}(\Tilde{\bm{g}}(\cdot)) + \mathcal{L}_{\text{Co}}(\Tilde{\bm{f}}(\cdot)) + \mathcal{L}_{\text{So}}(\Tilde{\bm{f}}(\cdot))],
\end{align}
\vspace{-15pt}

\noindent where each loss term corresponds to one of the four attributes of a dependable model:

\vspace{-15pt}
\begin{align}
\mathcal{L}_{\text{Ci}}(\Tilde{\bm{g}}(\cdot)) &= \mathbb{E}x [\mathbb{D}_1(\mathbf{T}_{k_1}(\Tilde{g}(x)),\mathbf{T}_{k_1}(g(x)))], \\
\mathcal{L}_{\text{Si}}(\Tilde{\bm{g}}(\cdot)) &= \mathbb{E}{x, \bm{\epsilon_1}} [\mathbb{D}_2(\mathbf{T}_{k_2}(\Tilde{g}(x)), \mathbf{T}_{k_2}(\Tilde{g}(x + \bm{\epsilon_1})))], \\
\mathcal{L}_{\text{Co}}(\Tilde{\bm{f}}(\cdot)) &= \mathbb{E}x [\mathbb{D}_3(\Tilde{f}(g(x)), f(g(x)))], \\
\mathcal{L}_{\text{So}}(\Tilde{\bm{f}}(\cdot)) &= \mathbb{E}{x, \bm{\epsilon_2}} [\mathbb{D}_4(\Tilde{f}(g(x)), \Tilde{f}(g(x + \bm{\epsilon_2})))],
\end{align}
\vspace{-15pt}

\noindent and $\mathbb{D}_i$ for $i \in \{1, 2, 3, 4\}$ are measures of probabilistic divergence, $\bm{\epsilon_i}$ is a perturbation with norm constrained by $\rho_i$ for $i \in \{1, 2\}$, $\mathbb{E}_{x, \bm{\epsilon_i}}$ denotes the expectation over both the input space and perturbation space and $\mathbf{T}_{k_i}(\cdot)$ for $i \in \{1,2\}$ denotes the top-$k$ function.

To achieve this, we incorporate the following loss terms into the optimization process:

\vspace{-15pt}
\begin{align}
\label{eq:allll}
\mathcal{L} = \mathcal{L}_{init} + \lambda_1 \cdot \mathcal{L}_{\text{Ci}} + \lambda_2 \cdot \mathcal{L}_{\text{Si}} + \lambda_3 \cdot \mathcal{L}_{\text{Co}} + \lambda_4 \cdot \mathcal{L}_{\text{So}},
\end{align}
\vspace{-15pt}

\noindent where $\mathcal{L}_{init}$ is the base loss function of DCG, and $\lambda_i$ for $i \in \{1, 2, 3, 4\}$ are regularization parameters that control the influence of each auxiliary loss term.

However, a major challenge is that the top-$k$ overlap functions for interpretable concept output $\mathbb{D}_1(\mathbf{T}_{k_1}(\Tilde{g}(x)),\mathbf{T}_k(g(x)))$ and $\mathbb{D}_2(\mathbf{T}_{k_2}(\Tilde{g}(x)), \mathbf{T}_k(\Tilde{g}(x + \bm{\epsilon_1})))$ are non-differentiable, which makes it impossible to use gradient descent. Therefore, we need to consider a surrogate loss for $\mathcal{L}_{\text{Ci}}(\Tilde{\bm{g}}(\cdot))$ and $\mathcal{L}_{\text{Si}}(\Tilde{\bm{g}}(\cdot))$.

To address this issue, we propose minimizing the distance between $\bm{\Tilde{g}(x)}$ and ${\bm{g(x)}}$ constrained on the top-$k$ entries only. Specifically, we minimize $||\bm{g(x)}_{S_{g(\cdot)}^{k_1}} - \bm{\Tilde{g}(x)}_{S_{\Tilde{g}(\cdot)}^{k_1}}||_1$, where $\bm{g(x)}_{S_{g(\cdot)}^{k_1}},\bm{\Tilde{g}(x)}_{S_{\Tilde{g}(\cdot)}^{k_1}} \in \mathbb{R}^k$ are the vectors $\bm{g(x)}$ and ${\bm{\Tilde{g}(x)}}$, respectively, constrained on the top-$k$ indices set $S_{g(\cdot)}^{k_1}$ of $g(x)$. Similarly, we minimize $||\bm{\Tilde{g}(x)}_{S_{\Tilde{g}(\cdot)}^{k_2}}-\bm{\Tilde{g}(x+\epsilon_1)}_{S_{\Tilde{g}(\cdot)}^{k_2}}||_1$, where $\bm{g(x+\epsilon_1)}_{S_{g(\cdot)}^{k_2}},\bm{\Tilde{g}(x+\epsilon_1)}_{S_{\Tilde{g}(\cdot)}^{k_2}} \in \mathbb{R}^k$ are the vectors $\bm{\Tilde{g}(x)}$ and $\bm{\Tilde{g}(x+\epsilon_1)}$, respectively, constrained on the top-$k$ indices set $S_{\Tilde{g}(\cdot)}^{k_2}$ of $\Tilde{g}(x+\epsilon_1)$. We use both top-$k$ indices sets for both vectors to involve the top-$k$ indices formation. Therefore, our surrogate loss functions are:

\begin{scriptsize}
\vspace{-13pt}
\begin{equation}
\label{eq:lcin}
\mathcal{L}_{\text{Ci}} = \frac{ ||\bm{g(x)}_{S_{g(\cdot)}^{k_1}} - \bm{\Tilde{g}(x)}_{S_{\Tilde{g}(\cdot)}^{k_1}}||_1 + || \bm{\Tilde{g}(x)}_{S_{\Tilde{g}(x)}^{k_1}} - \bm{g(x)}_{S_{g(x)}^{k_1}} ||_1}{2k_1}
\end{equation}
\vspace{-10pt}
\end{scriptsize}

\begin{scriptsize}
\vspace{-15pt}
\begin{equation}
\label{eq:sin}
\mathcal{L}_{\text{Si}} = \frac{ ||\bm{\Tilde{g}(x)}_{S_{\Tilde{g}(\cdot)}^{k_2}} - \bm{\Tilde{g}(x+\epsilon_1)}_{S_{\Tilde{g}(\cdot)}^{k_2}}||_1 + || \bm{\Tilde{g}(x+\epsilon_1)}_{S_{\Tilde{g}(\cdot)}^{k_2}} - \bm{\Tilde{g}(x)}_{S_{\Tilde{g}(\cdot)}^{k_2}} ||_1}{2k_2}
\end{equation}
\vspace{-10pt}
\end{scriptsize}

Inspired by the Projected Gradient Descent (PGD) method as described by Madry et al. \cite{madry2018towards}, we iteratively update the perturbation $\bm{\epsilon_i}$ (we define $\epsilon_1 = \epsilon_2 = \epsilon$ the sake of simplicity) and the model parameters ${\Tilde{f}(\cdot),\Tilde{g}(\cdot)}$ such that the model remains robust against adversarial perturbations while maintaining interpretability and predictive consistency. The iterative process involves the following steps for the $p$-th iteration. Mathematically,

\vspace{-15pt}
\begin{align}
\small
\bm{\epsilon_p} = \bm{\epsilon^*_{p-1}} + \frac{\alpha_p}{|A_{p-1}|} \sum_{x \in A_{p-1}} \nabla_{\bm{\delta^*_{p-1}}} \left( \mathcal{L}_{\text{Ci}} + \mathcal{L}_{\text{Si}} + \mathcal{L}_{\text{Co}} + \mathcal{L}_{\text{So}} \right),
\vspace{-15pt}
\end{align}
\vspace{-10pt}

\noindent where $\bm{\epsilon_p^{*}} = \arg\min_{\bm{||\epsilon||} \leq \max(\rho_1,\rho_2)} ||\bm{\epsilon - \epsilon_p}||$, $A_{p-1}$ is a batch of samples, $\alpha_p$ is the step size parameter for PGD.

After obtaining $\bm{\epsilon}_P$ after $P$ iterations, we update $\Tilde{f}^{t-1}(\cdot)$ and $\Tilde{g}^{t-1}(\cdot)$ to $\Tilde{f}^{t}(\cdot)$ and $\Tilde{g}^{t}(\cdot)$ using batched gradients. The final objective function ensures that the model adheres to the principles of consistent interpretability, stable interpretability, consistent output, and stable output, thereby achieving dependable performance across varying conditions.

\begin{table*}[htbp]
\vspace{-10pt}
  \centering
  \caption{Experimental results for DRIVE and baseline DCG. P1, P2, and P3 denote three different perturbations (see Section \ref{sec:per}).}
   \vspace{-10pt}
  \label{tab:1}
  \resizebox{\textwidth}{!}{
  \begin{threeparttable}
    \begin{tabular}{
      >{\centering\arraybackslash}m{2em} 
      *{3}{>{\centering\arraybackslash}m{4em}} 
      >{\centering\arraybackslash}p{10em} 
      *{3}{>{\centering\arraybackslash}m{4em}} 
      >{\centering\arraybackslash}p{10em} 
      *{3}{>{\centering\arraybackslash}m{4em}} 
      >{\centering\arraybackslash}p{10em} 
    }
    \toprule
        & \makecell{\textbf{a-MAE $\downarrow$}} & \makecell{\textbf{d-MAE $\downarrow$}} & \makecell{\textbf{(a,d)-MAE $\downarrow$}} & \makecell{\textbf{Top-k $\uparrow$}} 
        & \makecell{\textbf{a-MAE $\downarrow$}} & \makecell{\textbf{d-MAE $\downarrow$}} & \makecell{\textbf{(a,d)-MAE $\downarrow$}} & \makecell{\textbf{Top-k $\uparrow$}} 
        & \makecell{\textbf{a-MAE $\downarrow$}} & \makecell{\textbf{d-MAE $\downarrow$}} & \makecell{\textbf{(a,d)-MAE $\downarrow$}} & \makecell{\textbf{Top-k $\uparrow$}}  \\
        \midrule
    & \multicolumn{4}{c}{No} & \multicolumn{4}{c}{P1(0.08)} & \multicolumn{4}{c}{P3(0.01)} \\
    \cmidrule(lr){2-5} \cmidrule(lr){6-9} \cmidrule(lr){10-13}
    \textbf{DCG} & 2.336 & 4.593 & (2.339,6.1938) & \textbackslash{} & 2.5743 & 6.2599 & (2.6876,6.7961) & 0.336 & 2.614 & 19.701 & (2.6026,16.8892) & 0.301 \\
    \textbf{DRIVE} & \textbf{2.317} & \textbf{4.399} & \textbf{(2.291,6.0187)} & \textbackslash{} & \textbf{0.667} & \textbf{5.868} & \textbf{(2.0214,6.3694)} & \textbf{0.489} & \textbf{2.398} & \textbf{8.927} & \textbf{(2.307,8.972)} & \textbf{0.4425} \\
    \midrule
    & \multicolumn{4}{c}{P2(10\%)} & \multicolumn{4}{c}{P1(0.10)} & \multicolumn{4}{c}{P3(0.02)} \\
    \cmidrule(lr){2-5} \cmidrule(lr){6-9} \cmidrule(lr){10-13}
    \textbf{DCG} & 3.287 & 6.2457 & (3.411,8.1054) & 0.295 & 1.7519 & 6.7574 & (2.6852,14.0672) & 0.274 & 42.622 & 23.507 & (29.8753,45.1413) & 0.325 \\
    \textbf{DRIVE} & \textbf{2.5801} & \textbf{5.333} & \textbf{(3.278,5.291)} & \textbf{0.415} & \textbf{1.6182} & \textbf{5.7461} & \textbf{(2.3252,8.187)} & \textbf{0.4145} & \textbf{10.372} & \textbf{11.428} & \textbf{(11.3861,11.4262)} & \textbf{0.4532} \\
    \bottomrule
    \end{tabular}
    \end{threeparttable}}
\end{table*}

\begin{table}[htbp]
\vspace{-10pt}
  \centering
  \caption{Results of ablation experiments at P1 (0.08) by DRIVE.}
   \vspace{-10pt}
    \begin{tabular}{ccccccc}
    \toprule
    \multicolumn{1}{p{0.1em}}{\textbf{A}} & \multicolumn{1}{p{0.3em}}{\textbf{BC}} & \multicolumn{1}{p{0.3em}}{\textbf{DE}} & \multicolumn{1}{p{4em}}{\textbf{a-MAE $\downarrow$}} & \multicolumn{1}{p{4em}}{\textbf{d-MAE $\downarrow$}} & \multicolumn{1}{p{7em}}{\textbf{(a,d)-MAE $\downarrow$}} & \multicolumn{1}{p{3.5em}}{\textbf{Top-k $\uparrow$}} \\
    \midrule
    $\checkmark$     & $\times$     & $\times$    & 2.5743 & 6.2599 & (2.6876,6.7961) & 0.336 \\
    $\checkmark$     & $\checkmark$     & $\times$     & 1.9427 & 6.0311 & (2.4434,6.5846) & 0.472 \\
    $\checkmark$     & $\times$     & $\checkmark$     & 0.9291 & 5.9277 & (2.2778,6.4438) & 0.395 \\
    $\checkmark$     & $\checkmark$     & $\checkmark$     & \textbf{0.667} & \textbf{5.868} & \textbf{(2.0214,6.3694)} & \textbf{0.489} \\
    \bottomrule
    \end{tabular}%
    \vspace{-10pt}
  \label{tab:2}%
\end{table}%

\vspace{-5pt}
\section{EXPERIMENTS}
\vspace{-5pt}

Under various interference conditions, such as concept differences based on the output of large language models and the degradation of image quality, we have conducted a series of comparative experiments to demonstrate the significant improvement of our strategy in terms of output stability and robustness compared to the baseline (DCG) and non-CBM structured models. The experimental results will be presented through the model's predictive performance (measured by Mean Absolute Error, MAE) and the stability of the top K concepts in the intermediate concept layer.

\vspace{-5pt}
\subsection{Perturbations}
\label{sec:per}
\vspace{-5pt}
To evaluate the robustness and stability of the DRIVE model, we applied three distinct levels of noise perturbations:

\noindent\textbf{(P1) Gaussian Noise on Input Images.} Thermal noise is a prevalent form of image degradation. We introduced Gaussian noise with a mean \(\mu\) of 0 and a standard deviation \(\sigma\) of 0.03/0.08/0.10 to each frame of the real driving images in the test set, maintaining the frame size of \(224 \times 224\).

\noindent\textbf{(P2) Randomness in the Concept Set.} Concepts generated by large language models, such as GPT-3.5, exhibit inherent randomness and uncertainty, leading to non-identical concept data upon repeated generations. To simulate this, we created an additional conceptual dataset by replacing 5\%/10\%/15\% of the conceptual statements with synonyms. Training the model on this dataset revealed the impact of conceptual set instability on overall model performance.

\noindent\textbf{(P3) Perturbation to Model Parameters.} Noise in model parameters, though rare, can significantly degrade predictive capabilities. In our experiment, we added Gaussian noise with \(\mu = 0\) and \(\sigma = 0.01/0.02/0.03\) to all model parameters to assess the effect of such perturbations.

\vspace{-7pt}
\subsection{Dataset and Backbones:}

\noindent\textbf{Comma2k19.} This dataset, released by comma.ai, includes over 33 hours of commute data on California's Highway 280, divided into 2019 one-minute segments. Each segment covers a 20-kilometer stretch between San Jose and San Francisco. The dataset is partitioned into chunks, with each containing approximately 200 minutes of data. For our study, the first three chunks were used for training, validation, and testing. Data was preprocessed following Jessica Echterhoff et al.'s method \cite{Echterhoff_2024_WACV}, down-sampling from 20 to 4 frames per second to reduce redundancy. Each sequence is divided into 240 samples, resized to 224 × 224 pixels, and split into a 0.85/0.05/0.1 train/val/test distribution.

\noindent\textbf{CLIP.} Proposed by Radford et al. \cite{radford2021learning}, CLIP is a pretrained image-text encoder used in multi-modal tasks \cite{lai2023faithful,lai2024shared,lai2023multimodal}. It consists of an image encoder and a text encoder, trained through contrastive learning to understand semantic connections between images and text.

\noindent\textbf{VIT.} The Vision Transformer (ViT) serves as our experimental backbone, using a transformer encoder to process image patches extracted via a convolutional projection layer, facilitating the capture of global image features.

\noindent\textbf{LongFormer.} As a temporal encoder, LongFormer utilizes an efficient self-attention mechanism to process video frame sequences, capturing long-range temporal dependencies.

\vspace{-7pt}
\subsection{Evaluation Metrics}

\noindent\textbf{MAE.} Following DCG \cite{Echterhoff_2024_WACV}, we use the Mean Absolute Error (MAE) to evaluate the models' accuracy in predicting steering angles and forward distances in sequential scenes.

\noindent\textbf{Concepts Top-$k$ Overlap.} Concepts represent the rationale behind driving decisions. However, under input perturbations, these concepts may not accurately reflect real-world information. Therefore, we use the Top-$k$ concepts overlap, to measure the impact of perturbations.

\vspace{-7pt}
\subsection{Hyperparameters and Device}

For training the DCG model, we used 200 epochs, with an additional 40 epochs for the DRIVE model. The Adam optimizer was employed with a learning rate of 1e-5 and a weight decay of 1e-5. The batch size was set to 4 to balance computational efficiency and model performance. All models were trained on an Nvidia A40 GPU, utilizing CUDA version 12.2. For PGD, $eps = 0.08$ and $\alpha = 0.001$. The number of PGD iterations, which is the number of times the gradient updates are performed during the generation of adversarial samples. In this implementation, it is set to $5$. $\lambda_1, \lambda_2 = 1e+2$ and $\lambda_3, \lambda_4 = 1e-2$.

\vspace{-7pt}
\subsection{Performance Results}

Table \ref{tab:1} presents the test results of our proposed DRIVE framework compared to the baseline method DCG under both non-perturbed conditions and varying levels of perturbation. The metrics evaluated include angles MAE, distances MAE, angles\&distances MAE (multi-tasks) and concepts top-$k$ overlap.

It is evident from Table \ref{tab:1} that DRIVE outperforms DCG even in the original setting without any perturbation. Interestingly, our framework not only maintains but also enhances performance in the absence of additional disturbances, achieving SOTA results. This superior performance can be attributed to the intrinsic noise present in the task data, which our method effectively mitigates, thus improving generalization capabilities.

Moreover, Table \ref{tab:1} also illustrates the predictive regression task performance using MAE across three distinct perturbation levels. It is clear that DRIVE consistently exhibits better accuracy than the baseline approaches across all perturbation scenarios. This indicates that our method not only sustains stability but also achieves higher precision relative to the baselines. Overall, our approach demonstrates consistent high accuracy under all perturbation conditions, underscoring its robustness in maintaining precision while integrating high interpretability and stability across various datasets.

Additionally, Table \ref{tab:1} analyzes the interpretability output stability of DRIVE and the baseline methods under diverse perturbations, measured by the Top-K overlap of concepts. The table indicates that DRIVE shows greater stability in concept weights, with smaller differences in the top K concept matrices before and after perturbations. These findings suggest that DRIVE successfully combines interpretability with strong perturbation resilience, establishing itself as a faithful and robust model.

Collectively, the results presented in Table \ref{tab:1} substantiate the superior stability of DRIVE over baseline models. DRIVE successfully provides faithful explanations of autonomous driving decisions, positioning itself as a promising concept-based modeling approach.

\subsection{Ablation Study}

In our ablation studies, we conducted a comprehensive assessment of each component detailed in Equation (\ref{eq:allll}), with the goal of evaluating the significance and efficacy of each element in enhancing model performance. To accomplish this, we designated $\mathcal{L}_{init}$, the first term in Equation (\ref{eq:allll}), as our primary loss function. We then systematically evaluated various configurations by selectively omitting pairs of regularizers $\mathcal{L}_{Ci}$--B, $\mathcal{L}_{Si}$--C, $\mathcal{L}_{Co}$--D, and $\mathcal{L}_{So}$--E.

The results of our investigation (as shown in Table \ref{tab:2}) unequivocally demonstrate that each regularizer within our objective function is essential and effective. Each component makes a unique contribution to the enhancement of the model's performance. Of particular note is the inclusion of $\mathcal{L}_{Ci}$ and $\mathcal{L}_{Si}$, which significantly bolsters the stability of the model's concept outputs. And $\mathcal{L}_{Co}$ and $\mathcal{L}_{So}$ bolsters the stability and performance of the model's prediction outputs. This finding highlights the crucial role this regularization module plays in advancing model performance.

\section{CONCLUSIONS}

In conclusion, this study addresses the critical challenge of instability in end-to-end unsupervised explainable autonomous driving models, exemplified by DCG model. We introduce DRIVE — a Dependable, Robust, Interpretable, Visionary Ensemble Framework—to enhance the reliability and robustness of explanations and outputs. Through rigorous empirical evaluations, we demonstrate that DRIVE achieves consistent and stable interpretability and output, making significant strides toward more trustworthy autonomous driving systems. Our work lays the foundation for broader acceptance and deployment of autonomous vehicles by ensuring they meet stringent safety and regulatory standards. Future research could further refine DRIVE and explore its application in more diverse and challenging environments.

\footnotesize

\clearpage
\bibliography{ref}

\begin{thebibliography}{10}

\bibitem{yurtsever2020survey}
E.~Yurtsever, J.~Lambert, A.~Carballo, and K.~Takeda, ``A survey of autonomous driving: Common practices and emerging technologies,'' {\em IEEE access}, vol.~8, pp.~58443--58469, 2020.

\bibitem{zhao2023autonomous}
J.~Zhao, W.~Zhao, B.~Deng, Z.~Wang, F.~Zhang, W.~Zheng, W.~Cao, J.~Nan, Y.~Lian, and A.~F. Burke, ``Autonomous driving system: A comprehensive survey,'' {\em Expert Systems with Applications}, p.~122836, 2023.

\bibitem{li2023trustworthy}
B.~Li, P.~Qi, B.~Liu, S.~Di, J.~Liu, J.~Pei, J.~Yi, and B.~Zhou, ``Trustworthy ai: From principles to practices,'' {\em ACM Computing Surveys}, vol.~55, no.~9, pp.~1--46, 2023.

\bibitem{liao2024gpt}
H.~Liao, H.~Shen, Z.~Li, C.~Wang, G.~Li, Y.~Bie, and C.~Xu, ``Gpt-4 enhanced multimodal grounding for autonomous driving: Leveraging cross-modal attention with large language models,'' {\em Communications in Transportation Research}, vol.~4, p.~100116, 2024.

\bibitem{liao2024cognitive}
H.~Liao, Y.~Li, Z.~Li, C.~Wang, Z.~Cui, S.~E. Li, and C.~Xu, ``A cognitive-based trajectory prediction approach for autonomous driving,'' {\em IEEE Transactions on Intelligent Vehicles}, 2024.

\bibitem{molnar2020interpretable}
C.~Molnar, G.~Casalicchio, and B.~Bischl, ``Interpretable machine learning--a brief history, state-of-the-art and challenges,'' in {\em Joint European conference on machine learning and knowledge discovery in databases}, pp.~417--431, Springer, 2020.

\bibitem{wen2023road}
L.~Wen, X.~Yang, D.~Fu, X.~Wang, P.~Cai, X.~Li, T.~Ma, Y.~Li, L.~Xu, D.~Shang, {\em et~al.}, ``On the road with gpt-4v (ision): Early explorations of visual-language model on autonomous driving,'' {\em arXiv preprint arXiv:2311.05332}, 2023.

\bibitem{wang2024drive}
T.-H. Wang, A.~Maalouf, W.~Xiao, Y.~Ban, A.~Amini, G.~Rosman, S.~Karaman, and D.~Rus, ``Drive anywhere: Generalizable end-to-end autonomous driving with multi-modal foundation models,'' in {\em 2024 IEEE International Conference on Robotics and Automation (ICRA)}, pp.~6687--6694, IEEE, 2024.

\bibitem{liao2024human}
H.~Liao, S.~Liu, Y.~Li, Z.~Li, C.~Wang, Y.~Li, S.~E. Li, and C.~Xu, ``Human observation-inspired trajectory prediction for autonomous driving in mixed-autonomy traffic environments,'' in {\em 2024 IEEE International Conference on Robotics and Automation (ICRA)}, pp.~14212--14219, IEEE, 2024.

\bibitem{liao2024bat}
H.~Liao, Z.~Li, H.~Shen, W.~Zeng, D.~Liao, G.~Li, and C.~Xu, ``Bat: Behavior-aware human-like trajectory prediction for autonomous driving,'' in {\em Proceedings of the AAAI Conference on Artificial Intelligence}, vol.~38, pp.~10332--10340, 2024.

\bibitem{rudin2019stop}
C.~Rudin, ``Stop explaining black box machine learning models for high stakes decisions and use interpretable models instead,'' {\em Nature machine intelligence}, vol.~1, no.~5, pp.~206--215, 2019.

\bibitem{gdpr2016general}
G.~D. P.~R. GDPR, ``General data protection regulation,'' {\em Regulation (EU) 2016/679 of the European Parliament and of the Council of 27 April 2016 on the protection of natural persons with regard to the processing of personal data and on the free movement of such data, and repealing Directive 95/46/EC}, 2016.

\bibitem{gdpr2022automated}
E.~GDPR, ``Automated individual decision-making, including profiling,'' 2022.

\bibitem{omeiza2021explanations}
D.~Omeiza, H.~Webb, M.~Jirotka, and L.~Kunze, ``Explanations in autonomous driving: A survey,'' {\em IEEE Transactions on Intelligent Transportation Systems}, vol.~23, no.~8, pp.~10142--10162, 2021.

\bibitem{atakishiyev2024explainable}
S.~Atakishiyev, M.~Salameh, H.~Yao, and R.~Goebel, ``Explainable artificial intelligence for autonomous driving: A comprehensive overview and field guide for future research directions,'' {\em IEEE Access}, 2024.

\bibitem{2022Explainability}
Z.~Loi, B.-Y. Hédi, P.~Patrick, and C.~Matthieu, ``Explainability of deep vision-based autonomous driving systems: Review and challenges,'' {\em International Journal of Computer Vision}, 2022.

\bibitem{8237582}
J.~Kim and J.~Canny, ``Interpretable learning for self-driving cars by visualizing causal attention,'' in {\em 2017 IEEE International Conference on Computer Vision (ICCV)}, pp.~2961--2969, 2017.

\bibitem{9710855}
K.~Chitta, A.~Prakash, and A.~Geiger, ``Neat: Neural attention fields for end-to-end autonomous driving,'' in {\em 2021 IEEE/CVF International Conference on Computer Vision (ICCV)}, pp.~15773--15783, 2021.

\bibitem{9157111}
Y.~Xu, X.~Yang, L.~Gong, H.-C. Lin, T.-Y. Wu, Y.~Li, and N.~Vasconcelos, ``Explainable object-induced action decision for autonomous vehicles,'' in {\em 2020 IEEE/CVF Conference on Computer Vision and Pattern Recognition (CVPR)}, pp.~9520--9529, 2020.

\bibitem{8954347}
W.~Zeng, W.~Luo, S.~Suo, A.~Sadat, B.~Yang, S.~Casas, and R.~Urtasun, ``End-to-end interpretable neural motion planner,'' in {\em 2019 IEEE/CVF Conference on Computer Vision and Pattern Recognition (CVPR)}, pp.~8652--8661, 2019.

\bibitem{Echterhoff_2024_WACV}
J.~Echterhoff, A.~Yan, K.~Han, A.~Abdelraouf, R.~Gupta, and J.~McAuley, ``Driving through the concept gridlock: Unraveling explainability bottlenecks in automated driving,'' in {\em Proceedings of the IEEE/CVF Winter Conference on Applications of Computer Vision (WACV)}, pp.~7346--7355, January 2024.

\bibitem{10287869}
Z.~Zheng, Y.~Cheng, Z.~Xin, Z.~Yu, and B.~Zheng, ``Robust perception under adverse conditions for autonomous driving based on data augmentation,'' {\em IEEE Transactions on Intelligent Transportation Systems}, vol.~24, no.~12, pp.~13916--13929, 2023.

\bibitem{10009752}
Z.~Li, ``Lidar-based 3d object detection for autonomous driving,'' in {\em 2022 International Conference on Image Processing, Computer Vision and Machine Learning (ICICML)}, pp.~507--512, 2022.

\bibitem{Almalioglu2022}
Y.~Almalioglu, M.~Turan, N.~Trigoni, and A.~Markham, ``Deep learning-based robust positioning for all-weather autonomous driving,'' {\em Nature Machine Intelligence}, vol.~4, pp.~749--760, Sep 2022.

\bibitem{oikarinenlabel}
T.~Oikarinen, S.~Das, L.~M. Nguyen, and T.-W. Weng, ``Label-free concept bottleneck models,'' in {\em International Conference on Learning Representations}, 2023.

\bibitem{lai2023faithful}
S.~Lai, L.~Hu, J.~Wang, L.~Berti-Equille, and D.~Wang, ``Faithful vision-language interpretation via concept bottleneck models,'' in {\em The Twelfth International Conference on Learning Representations}, 2023.

\bibitem{hu2023seat}
L.~Hu, Y.~Liu, N.~Liu, M.~Huai, L.~Sun, and D.~Wang, ``Seat: stable and explainable attention,'' in {\em Proceedings of the AAAI Conference on Artificial Intelligence}, vol.~37, pp.~12907--12915, 2023.

\bibitem{wiegreffe2019attention}
S.~Wiegreffe and Y.~Pinter, ``Attention is not not explanation,'' in {\em Proceedings of the 2019 Conference on Empirical Methods in Natural Language Processing and the 9th International Joint Conference on Natural Language Processing (EMNLP-IJCNLP)}, pp.~11--20, 2019.

\bibitem{jacovi2020towards}
A.~Jacovi and Y.~Goldberg, ``Towards faithfully interpretable nlp systems: How should we define and evaluate faithfulness?,'' in {\em Proceedings of the 58th Annual Meeting of the Association for Computational Linguistics}, pp.~4198--4205, 2020.

\bibitem{lyu2022towards}
Q.~Lyu, M.~Apidianaki, and C.~Callison-Burch, ``Towards faithful model explanation in nlp: A survey,'' {\em arXiv preprint arXiv:2209.11326}, 2022.

\bibitem{yeh2019fidelity}
C.~Yeh, C.~Hsieh, A.~S. Suggala, D.~I. Inouye, and P.~Ravikumar, ``On the (in)fidelity and sensitivity of explanations,'' in {\em Advances in Neural Information Processing Systems 32: Annual Conference on Neural Information Processing Systems 2019, NeurIPS 2019, December 8-14, 2019, Vancouver, BC, Canada}, pp.~10965--10976, 2019.

\bibitem{adebayo2018sanity}
J.~Adebayo, J.~Gilmer, M.~Muelly, I.~Goodfellow, M.~Hardt, and B.~Kim, ``Sanity checks for saliency maps,'' {\em Advances in neural information processing systems}, vol.~31, 2018.

\bibitem{yin2022sensitivity}
F.~Yin, Z.~Shi, C.-J. Hsieh, and K.-W. Chang, ``On the sensitivity and stability of model interpretations in nlp,'' in {\em Proceedings of the 60th Annual Meeting of the Association for Computational Linguistics (Volume 1: Long Papers)}, pp.~2631--2647, 2022.

\bibitem{lai2024fts}
S.~Lai, N.~Feng, H.~Sui, Z.~Ma, H.~Wang, Z.~Song, H.~Zhao, and Y.~Yue, ``Fts: A framework to find a faithful timesieve,'' {\em arXiv preprint arXiv:2405.19647}, 2024.

\bibitem{Dodge_2019}
J.~Dodge, Q.~V. Liao, Y.~Zhang, R.~K.~E. Bellamy, and C.~Dugan, ``Explaining models: an empirical study of how explanations impact fairness judgment,'' in {\em Proceedings of the 24th International Conference on Intelligent User Interfaces}, IUI ’19, ACM, Mar. 2019.

\bibitem{DING2024120057}
W.~Ding, I.~Alrashdi, H.~Hawash, and M.~Abdel-Basset, ``Deepsecdrive: An explainable deep learning framework for real-time detection of cyberattack in in-vehicle networks,'' {\em Information Sciences}, vol.~658, p.~120057, 2024.

\bibitem{koh2020concept}
P.~W. Koh, T.~Nguyen, Y.~S. Tang, S.~Mussmann, E.~Pierson, B.~Kim, and P.~Liang, ``Concept bottleneck models,'' in {\em International Conference on Machine Learning}, pp.~5338--5348, PMLR, 2020.

\bibitem{oikarinen2023label}
T.~Oikarinen, S.~Das, L.~M. Nguyen, and T.-W. Weng, ``Label-free concept bottleneck models,'' {\em arXiv preprint arXiv:2304.06129}, 2023.

\bibitem{neimark2021video}
D.~Neimark, O.~Bar, M.~Zohar, and D.~Asselmann, ``Video transformer network,'' in {\em Proceedings of the IEEE/CVF International Conference on Computer Vision}, pp.~3163--3172, 2021.

\bibitem{vivit}
A.~Arnab, M.~Dehghani, G.~Heigold, C.~Sun, M.~Lucic, and C.~Schmid, ``Vivit: {A} video vision transformer,'' {\em CoRR}, vol.~abs/2103.15691, 2021.

\bibitem{ouyang2022training}
L.~Ouyang, J.~Wu, X.~Jiang, D.~Almeida, C.~Wainwright, P.~Mishkin, C.~Zhang, S.~Agarwal, K.~Slama, A.~Ray, {\em et~al.}, ``Training language models to follow instructions with human feedback,'' {\em Advances in Neural Information Processing Systems}, vol.~35, pp.~27730--27744, 2022.

\bibitem{caesar2020nuscenes}
H.~Caesar, V.~Bankiti, A.~H. Lang, S.~Vora, V.~E. Liong, Q.~Xu, A.~Krishnan, Y.~Pan, G.~Baldan, and O.~Beijbom, ``nuscenes: A multimodal dataset for autonomous driving,'' in {\em Proceedings of the IEEE/CVF conference on computer vision and pattern recognition}, pp.~11621--11631, 2020.

\bibitem{radford2021learning}
A.~Radford, J.~W. Kim, C.~Hallacy, A.~Ramesh, G.~Goh, S.~Agarwal, G.~Sastry, A.~Askell, P.~Mishkin, J.~Clark, {\em et~al.}, ``Learning transferable visual models from natural language supervision,'' in {\em International conference on machine learning}, pp.~8748--8763, PMLR, 2021.

\bibitem{beltagy2020longformer}
I.~Beltagy, M.~E. Peters, and A.~Cohan, ``Longformer: The long-document transformer,'' {\em arXiv preprint arXiv:2004.05150}, 2020.

\bibitem{DBLP:journals/corr/HendrycksG16}
D.~Hendrycks and K.~Gimpel, ``Bridging nonlinearities and stochastic regularizers with gaussian error linear units,'' {\em CoRR}, vol.~abs/1606.08415, 2016.

\bibitem{srivastava2014dropout}
N.~Srivastava, G.~Hinton, A.~Krizhevsky, I.~Sutskever, and R.~Salakhutdinov, ``Dropout: a simple way to prevent neural networks from overfitting,'' {\em The journal of machine learning research}, vol.~15, no.~1, pp.~1929--1958, 2014.

\bibitem{madry2018towards}
A.~Madry, A.~Makelov, L.~Schmidt, D.~Tsipras, and A.~Vladu, ``Towards deep learning models resistant to adversarial attacks,'' in {\em International Conference on Learning Representations}, 2018.

\bibitem{lai2024shared}
S.~Lai, J.~Li, G.~Guo, X.~Hu, Y.~Li, Y.~Tan, Z.~Song, Y.~Liu, Z.~Ren, C.~Wang, {\em et~al.}, ``Shared and private information learning in multimodal sentiment analysis with deep modal alignment and self-supervised multi-task learning,'' in {\em 2024 International Joint Conference on Neural Networks (IJCNN)}, pp.~1--8, IEEE, 2024.

\bibitem{lai2023multimodal}
S.~Lai, X.~Hu, H.~Xu, Z.~Ren, and Z.~Liu, ``Multimodal sentiment analysis: A survey,'' {\em Displays}, p.~102563, 2023.

\end{thebibliography}
\bibliographystyle{ieeetr}

\end{document}